# Crash Severity Analysis of Child Bicyclists using Arm-Net and MambaNet

Shriyank Somvanshi
*Ingram School of Engineering, Texas State University*
601 University Drive, San Marcos, Texas 78666
shriyank@txstate.edu

Rohit Chakraborty
*Ingram School of Engineering, Texas State University*
601 University Drive, San Marcos, Texas 78666
xuw12@txstate.edu

Subasish Das
*Ingram School of Engineering, Texas State University*
601 University Drive, San Marcos, Texas 78666
subasish@txstate.edu

Anandi K Dutta
*Ingram School of Engineering, Texas State University*
601 University Drive, San Marcos, Texas 78666
myb43@txstate.edu

*Abstract*—Child bicyclists (14 years and younger) are among the most vulnerable road users, often experiencing severe injuries or fatalities in crashes. This study analyzed 2,394 child bicyclist crashes in Texas from 2017 to 2022 using two deep tabular learning models (ARM-Net and MambaNet). To address the issue of data imbalance, the SMOTEENN technique was applied, resulting in balanced datasets that facilitated accurate crash severity predictions across three categories: Fatal/Severe (KA), Moderate/Minor (BC), and No Injury (O). The findings revealed that MambaNet outperformed ARM-Net, achieving higher precision, recall, F1-scores, and accuracy, particularly in the KA and O categories. Both models highlighted challenges in distinguishing BC crashes due to overlapping characteristics. These insights underscored the value of advanced tabular deep learning methods and balanced datasets in understanding crash severity. While limitations such as reliance on categorical data exist, future research could explore continuous variables and real-time behavioral data to enhance predictive modeling and crash mitigation strategies.

## I. INTRODUCTION

Child bicyclists (14 years and younger) are among the most vulnerable road users, facing significant crash risks due to their limited experience, difficulty in assessing traffic situations, and reduced visibility to drivers [1]. These crashes frequently occur in complex environments like intersections and urban areas, where interactions with vehicles are unpredictable. Despite safety efforts, crashes involving child bicyclists remain high, often resulting in severe injuries or fatalities. Bicyclists represent about 2% of annual motor vehicle crash fatalities in the United States, with younger bicyclists contributing significantly [2]. From 2015 to 2022, male child bicyclists accounted for 6% to 11% of fatalities annually, compared to 1% to 2% for females [2]. In 2021, child bicyclists made up 4% of all fatalities and 12% of all injury crashes [1]. Addressing these risks is crucial not only for reducing injuries and fatalities but also for encouraging safe and sustainable cycling among children.

This study aimed to explore the factors associated with child bicyclist crashes and identify the crash patterns. To address the research gap in understanding factors influencing crash severity, this study employed ArmNet and MambaNet, along with the Synthetic Minority Oversampling Technique with Edited Nearest Neighbors (SMOTEENN) resampling technique to handle data imbalance. These methods allow for an in-depth exploration of crash attributes such as road conditions, demographics, and environmental factors. The study identifies critical risk factors by leveraging these models and provides actionable recommendations to improve infrastructure, safety policies, and child bicyclist safety.

## II. LITERATURE REVIEW

Child bicyclist safety is a key area in road safety research, with studies highlighting crash factors, injury risks, and potential interventions. Infrastructure plays a crucial role, as intersections and roads without cycling facilities pose significant hazards. Wang et al. [3] found that stop-controlled intersections and helmet use reduced injury severity, while uncontrolled intersections and adverse weather heightened risks. Vansteenkiste et al. [4] observed that child bicyclists exhibit reduced horizontal gaze on varying road surfaces, making them more vulnerable in complex traffic environments.

Children's impulsive behavior, rule-breaking, and limited hazard perception increase their crash risk, particularly in complex traffic situations. Liu et al. [5] found that inexperience and poor hazard recognition led to delayed or absent braking in response to right-turning motorcycles, heightening crash risks. Wang et al. [6] identified gender, age, and perceived risk as key factors, with younger cyclists and males more likely to take risks, such as violating traffic rules. Environmental factors also play a major role, as poor visibility and adverse weather conditions contribute to crash severity. Asgarzadeh et al. [7]reported that low-light conditions and wet roads significantly increased injury severity, while Wang et al. [3] found similar effects at unsignalized intersections.

Motor vehicle interactions and cycling infrastructure significantly impact child bicyclist crash risks, especially at intersections and in mixed traffic. Doong and Lai [8] found that right-turning vehicles and signal-free intersections posed the greatest risks in Taiwan, while Haworth et al. [9] noted that younger drivers were more likely to misjudge bicyclist movements in low-visibility conditions. Infrastructure also plays a critical role, as Gitelman et al. [10] found that young e-cyclists in Israel traveled at higher speeds, increasing crash risks in mixed-use urban areas. Paridon et al. [11] observed that UK child bicyclists struggled to detect hidden hazards compared to adults, emphasizing cognitive limitations that heighten crash risks.

Tabular deep learning has been widely used in traffic safety analysis, leveraging advanced architectures for interpretability and efficiency [12]. Studies have explored pedestrian crash severity using TabNet [13], while Hijazi et al. [14] evaluated multiple deep learning models, including TabNet, for crash severity prediction. To analyze child bicyclist crash patterns, this study employs ARM-Net [15] and MambaNet [16] to enhance severity prediction.

## III. DATA PREPARATION

This study analyzed 2,394 child bicyclist crashes in Texas (2017–2022) to examine crash severity across Fatal/Severe (KA), Moderate/Minor (BC), and No Injury (O) categories. **Fig. 1** shows that BC crashes peaked in 2019 before declining, likely due to safety improvements, while KA crashes remained stable, indicating persistent risks. O crashes peaked in 2018 and gradually decreased, suggesting better infrastructure and awareness. Feature importance analysis using XGBoost and Random Forest identified the top 10 common predictors, including weather, lighting, road alignment, surface conditions, traffic control, intersection involvement, first harmful event, road class, speed limit, and helmet use. Despite an overall decline, severe crashes still necessitate targeted safety interventions.

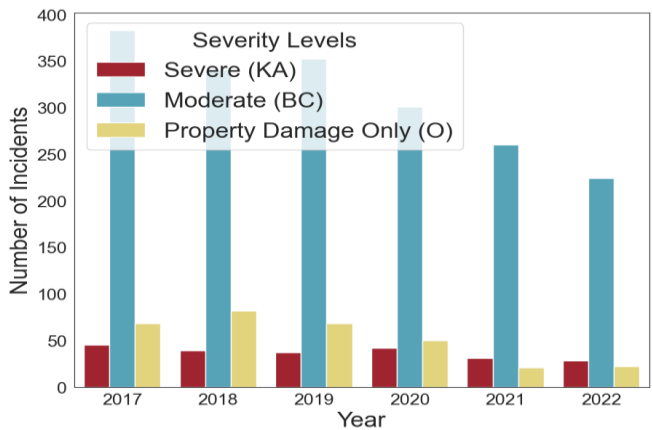

Fig. 1 Crash Severity Distribution by Year.

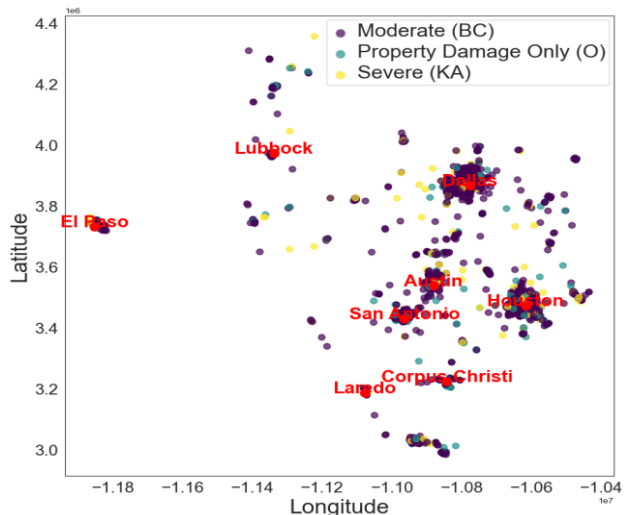

Fig. 2 Crash Locations by Severity Types.

**Fig. 2** shows that crashes are most concentrated in Dallas, Houston, Austin, and San Antonio, with BC crashes more prevalent in urban areas due to infrastructure and traffic risks. In contrast, KA and O crashes are more dispersed, while smaller cities like Lubbock and El Paso report fewer incidents, likely due to lower exposure. These findings highlight the need for targeted safety measures in high-risk urban areas.

## IV. METHODOLOGY

### A. Study Design

This study predicts crash severity using ARM-Net and MambaNet while addressing data imbalance with SMOTEENN. The methodology follows a structured workflow: Stage 1 involves data preprocessing, including feature scaling, encoding, and resampling through oversampling and noise reduction techniques. Stage 2 focuses on severity prediction, with the dataset split into 60% training, 20% validation, and 20% testing. Model performance is evaluated using accuracy, precision, recall, F1-score, and confusion matrix analysis. **Fig. 3** illustrates the two-stage process, highlighting the effectiveness of resampling in balancing class distributions and improving predictive performance.

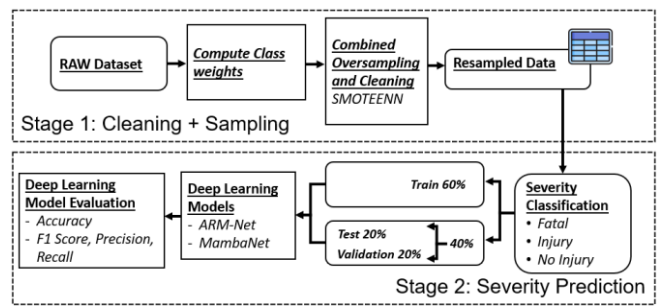

Fig. 3 Flowchart of the Combined Study.

This study predicts child bicyclist crash severity using ARM-Net and MambaNet. ARM-Net leverages exponential feature transformation and sparse multi-head gated attention to model feature interactions while filtering noise, improving accuracy and interpretability [17]. MambaNet combines convolutional layers and LSTM units to capture spatial and sequential dependencies in crash data, enhancing predictive performance [18]. These models were chosen for their ability to identify complex patterns in structured data. Preprocessing included standardizing feature names, distinguishing categorical and numerical variables, and applying normalization (StandardScaler) and one-hot encoding (OneHotEncoder) using scikit-learn.

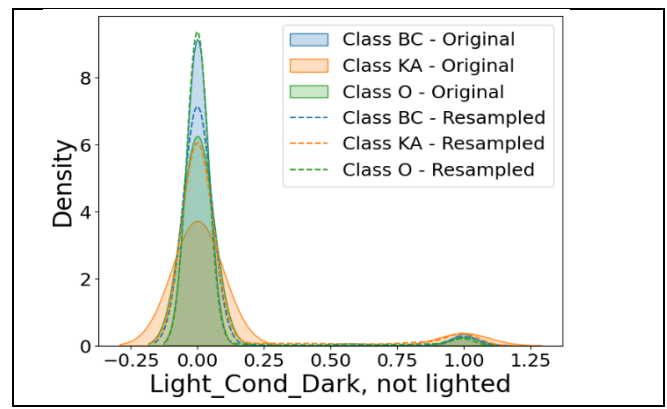

Fig. 4 Before and after feature distribution analysis with SMOTEENN, Class-conditional feature distributions of dark-not lighted condition

To address data imbalance, SMOTEENN, a hybrid resampling technique combining SMOTE and Edited Nearest

Neighbors (ENN), was applied to oversample minority classes and remove noisy samples. The resampled dataset includes 3,567 crashes, with 1,558 fatal, 520 injury, and 1,489 non-injury crashes, ensuring a balanced distribution. **Fig. 4** illustrates the feature distribution analysis, showing that resampling preserves original data characteristics, particularly in class-conditional distributions for 'Dark not lighted.' These results confirm SMOTEENN's effectiveness in improving class balance while maintaining data integrity, and enhancing crash severity prediction.

### B. Hyperparameter Tuning

Hyperparameter optimization is crucial for maximizing the performance of ARM-Net and MambaNet, as their effectiveness depends on selecting optimal parameters. To address variability, we implemented random search as recommended by Bergstra et al. [20], efficiently exploring the hyperparameter space for both models. Table I details the search space, which included 100 randomly sampled configurations per model, trained over 50 epochs using GPU acceleration. ARM-Net utilizes attention, memory, and residual components with key parameters such as hidden_dim: [128], num_layers: [4], and dropout_rate: [0.3], ensuring effective feature representation. MambaNet applies a multi-layered architecture and dropout strategies to enhance generalization, using hidden_dims: [128, 64], dropout_rate: [0.3], and weight_decay: [1e-4]. This systematic tuning approach improves model reliability and generalization.

Table I HYPERPARAMETER TUNING

| Model | Table Column Head |
|---|---|
| ARM-Net | input_dim: X_train.shape[1], hidden_dim: [128], output_dim: len(y_train.unique()), num_layers: [4], dropout_rate: [0.3], lr: [1e-3], weight_decay: [1e-4], epochs: [50], batch_size: [32], optimizer: AdamW, scheduler: ReduceLROnPlateau |
| MambaNet | input_dim: X_train.shape[1], hidden_dims: [128, 64], output_dim: len(label_encoder.classes_), dropout_rate: [0.3], lr: [1e-3], weight_decay: [1e-4], epochs: [50], batch_size: [32], optimizer: AdamW, scheduler: ReduceLROnPlateau |

## V. RESULTS AND DISCUSSION

### A. Validation of Experiment

**Table II** summarizes the training and validation performance of ARM-Net and MambaNet in predicting crash severity across KA, BC, and O categories. MambaNet achieved a higher accuracy (92%) than ARM-Net (88%), demonstrating superior predictive performance. Both models were trained on a balanced dataset, with 1,558 KA, 520 BC, and 1,489 O crashes, ensuring effective handling of imbalanced severity data. These results highlight the robustness of deep learning models in crash severity prediction.

Table II SUMMARY OF TRAINING AND VALIDATION FOR CRASH SEVERITY PREDICTION MODELS

| Model | Accuracy (%) | Epochs | Number of Samples | | |
|---|---|---|---|---|---|
| | | | KA | BC | O |
| ARM-Net | 88 | 50 (Early Stopping) | 1,558 | 520 | 1,489 |
| MambaNet | 92 | | 1,558 | 520 | 1,489 |

### B. Model Performance

**Table III** presents the prediction performance of ARM-Net and MambaNet across KA, BC, and O categories using precision, recall, F1-score, and accuracy metrics. ARM-Net performed well in the KA category, achieving 86% precision, 96% recall, and a 91% F1-score, with an overall accuracy of 95.63%, highlighting its reliability in identifying severe crashes. However, its performance declined in the BC category, with 87% precision, 67% recall, and a 76% F1-score, resulting in a lower accuracy of 66.67%. In the O category, it performed better, achieving 90% precision, 87% recall, and an 89% F1-score, with 87.21% accuracy. These results suggest that ARM-Net is effective in identifying extreme crash severity levels but struggles with moderate-severity cases.

MambaNet outperformed ARM-Net, particularly in the KA and O categories. It achieved 90% precision and 98% recall for KA crashes, with an F1-score of 94% and 97.78% accuracy, demonstrating strong reliability in identifying severe crashes. In the O category, it maintained 94% precision, recall, and F1-score, with an accuracy of 94.75%, highlighting its consistency in predicting non-injury crashes. However, like ARM-Net, MambaNet struggled with the BC category, achieving 90% precision but only 63% recall, leading to a 74% F1-score and 66.02% accuracy, indicating challenges in distinguishing moderate-severity crashes.

MambaNet outperformed ARM-Net, particularly in extreme severity categories (KA and O), achieving higher precision, recall, F1-scores, and accuracy. Its advanced architecture makes it more effective for identifying high-risk or no-injury crashes. However, both models struggled in the BC category, showing similar accuracy but lower recall, likely due to overlapping crash characteristics.

Table III PREDICTION PERFORMANCE OF ARM-NET AND MAMBANET MODELS

| Model | Category | Precision (%) | Recall (%) | F-1 Score(%) | Accuracy (%) |
|---|---|---|---|---|---|
| ARM-Net | KA | 86 | 96 | 91 | 95.63 |
| | BC | 87 | 67 | 76 | 66.67 |
| | O | 90 | 87 | 89 | 87.21 |
| MambaNet | KA | 90 | 98 | 94 | 97.78 |
| | BC | 90 | 63 | 74 | 66.02 |
| | O | 94 | 94 | 94 | 94.75 |

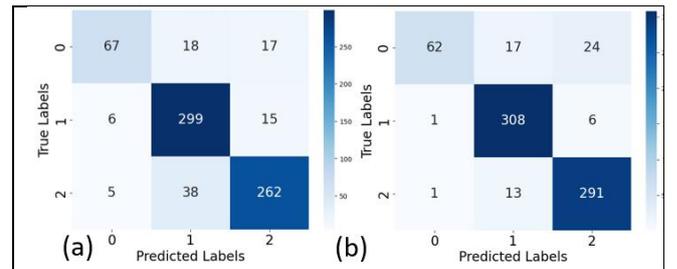

Fig. 5 Confusion Matrix of each model: (a) ARM-Net; (b) MambaNet

**Fig. 5** compares the classification performance of ARM-Net and MambaNet using confusion matrices. ARM-Net (**Fig. 5a**) misclassifies several O crashes as BC, affecting its accuracy. MambaNet (**Fig. 5b**) shows improved performance with higher true positive rates and fewer misclassifications, particularly in KA and O categories, demonstrating its superior ability to distinguish crash severity levels.

**Fig. 6** illustrates the log loss curves for ARM-Net and MambaNet. ARM-Net (**Fig. 6a**) initially learns effectively, but its validation loss fluctuates after 15 epochs, indicating potential overfitting. MambaNet (**Fig. 6b**) shows a smoother and more consistent reduction in loss over 40 epochs, with validation loss closely following training loss, suggesting better generalization and stability.

Integrating ARM-Net and MambaNet into real-time traffic safety systems could enhance proactive crash mitigation by leveraging IoT sensors, vehicle telematics, and surveillance data. MambaNet's superior performance in predicting severe and no-injury crashes makes it ideal for ITS and connected vehicle frameworks, enabling adaptive traffic control, automated safety alerts, and predictive road maintenance. Future research should explore hybrid approaches incorporating spatiotemporal crash data and real-time behavioral insights to improve predictive accuracy and intervention strategies.

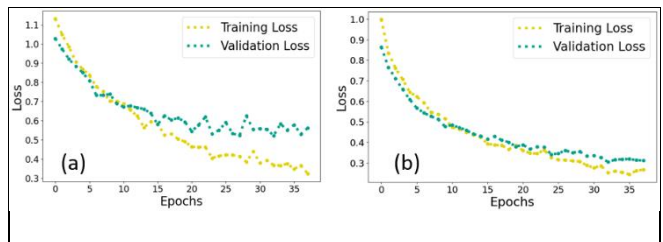

Fig. 6 Log loss of each model: (a) ARM-Net; (b) MambaNet

## VI. CONCLUSIONS

This study analyzed 2,394 child bicyclist crashes in Texas (2017–2022) using ARM-Net and MambaNet, with SMOTEENN applied to address data imbalance and improve crash severity predictions. The methodology included preprocessing, hyperparameter tuning, and evaluation metrics to ensure model robustness. Findings highlight the need for targeted safety interventions based on feature importance analysis. Key factors such as lighting, road alignment, traffic control, and surface conditions suggest practical measures like improved street lighting, better signage, and protected bike lanes. Behavioral factors like helmet use and speed limits emphasize the importance of enforcement campaigns and safety education programs to reduce crash severity.

Effective policy measures require better data collection and the integration of real-time crash monitoring technologies to support data-driven infrastructure improvements and enforcement strategies. Addressing data imbalance using methods like SMOTEENN can help policymakers develop equitable safety interventions for high-risk crash scenarios.

This study has some limitations, including reliance on categorical crash data, limiting insights into vehicle speeds and real-time behaviors. Future research should incorporate continuous variables, naturalistic data (e.g., video recordings, sensor-based observations), and simulation-based modeling to enhance understanding of child bicyclist safety across different cultural and regional contexts.

## REFERENCES

[1] NHTSA, "Traffic Safety Facts 2021 Data: Bicyclists and Other Cyclists," DOT HS 813 484, Jun. 2023. Accessed: Dec. 27, 2024. [Online]. Available: https://rosap.ntl.bts.gov/view/dot/78168
[2] IIHS, "Fatality Facts 2022: Bicyclists," IIHS-HLDI crash testing and highway safety. Accessed: Dec. 15, 2024. [Online]. Available: https://www.iihs.org/topics/fatality-statistics/detail/bicyclists
[3] C. Wang, L. Lu, and J. Lu, "Statistical Analysis of Bicyclists' Injury Severity at Unsignalized Intersections," *Traffic Injury Prevention*, vol. 16, no. 5, pp. 507–512, Jul. 2015, doi: 10.1080/15389588.2014.969802.
[4] P. Vansteenkiste, L. Zeuwts, M. van Maarseveen, G. Cardon, G. Savelsbergh, and M. Lenoir, "The Implications of Low Quality Bicycle Paths on the Gaze Behaviour of Young Learner Cyclists," *Transportation Research Part F: Traffic Psychology and Behaviour*, vol. 48, p. pp 52-60, Jul. 2017, doi: 10.1016/j.trf.2017.04.013.
[5] W.-C. Liu, M.-C. Jeng, J.-R. Hwang, J.-L. Doong, C.-Y. Lin, and C.-H. Lai, "The Response Patterns of Young Bicyclists to a Right-Turning Vehicle: A Simulator Study," *Perceptual and Motor Skills*, vol. 115, no. 2, p. pp 385-402, Oct. 2012, doi: 10.2466/22.25.27.PMS.115.5.385-402.
[6] C. Wang, W. Zhang, Z. Feng, K. Wang, and Y. Gao, "Exploring Factors Influencing the Risky Cycling Behaviors of Young Cyclists Aged 15–24 Years: A Questionnaire-Based Study in China," *Risk Analysis*, vol. 40, no. 8, p. pp 1554-1570, Aug. 2020, doi: 10.1111/risa.13499.
[7] M. Asgarzadeh, D. Fischer, S. K. Verma, T. K. Courtney, and D. C. Christiani, "The impact of weather, road surface, time-of-day, and light conditions on severity of bicycle-motor vehicle crash injuries," *American Journal of Industrial Medicine*, vol. 61, no. 7, pp. 556–565, 2018, doi: 10.1002/ajim.22849.
[8] J.-L. Doong and C.-H. Lai, "Risk Factors for Child and Adolescent Occupants, Bicyclists, and Pedestrians in Motorized Vehicle Collisions," *Traffic Injury Prevention*, vol. 13, no. 3, p. pp 249-257, May 2012, doi: 10.1080/15389588.2011.647140.
[9] N. Haworth, M. Legge, D. Twisk, J. Bonham, T. O'Hare, and M. Johnson, "Young Driver Crashes with Cyclists: Identifying Training Opportunities," *Transportation Research Record: Journal of the Transportation Research Board*, vol. 2673, no. 12, p. pp 679-689, Dec. 2019, doi: 10.1177/0361198119860118.
[10] V. GITELMAN, A. Korchatov, and W. Elias, "Speeds of young e-cyclists on urban streets and related risk factors: an observational study in Israel," *Safety*, vol. 6, no. 2, p. Article-ID 29, Jun. 2020, doi: 10.3390/safety6020029.
[11] K. N. van Paridon, H. K. Leivers, P. J. Robertson, and M. A. Timmis, "Visual search behaviour in young cyclists: A naturalistic experiment," *Transportation Research Part F: Traffic Psychology and Behaviour*, vol. 67, p. pp 217-229, Nov. 2019, doi: 10.1016/j.trf.2019.10.014.
[12] S. Somvanshi, S. Das, S. A. Javed, G. Antariksa, and A. Hossain, "A Survey on Deep Tabular Learning," Oct. 15, 2024, *arXiv*: arXiv:2410.12034. doi: 10.48550/arXiv.2410.12034.
[13] A. Rafe and P. A. Singleton, "Exploring Factors Affecting Pedestrian Crash Severity Using TabNet: A Deep Learning Approach," *Data Sci. Transp.*, vol. 6, no. 2, p. 13, Aug. 2024, doi: 10.1007/s42421-024-00098-x.
[14] H. Hijazi, K. Sattar, H. M. Al-Ahmadi, and S. El-Ferik, "Comparative Study for Optimized Deep Learning-Based Road Accidents Severity Prediction Models," *Arab J Sci Eng*, vol. 49, no. 4, pp. 5853–5873, Apr. 2024, doi: 10.1007/s13369-023-08510-4.
[15] S. Cai, K. Zheng, G. Chen, H. V. Jagadish, B. C. Ooi, and M. Zhang, "ARM-Net: Adaptive Relation Modeling Network for Structured Data," in *Proceedings of the 2021 International Conference on Management of Data*, Jun. 2021, pp. 207–220. doi: 10.1145/3448016.3457321.
[16] R. Khanmohammadi, S. Saba-Sadiya, S. Esfandiarpour, T. Alhanai, and M. M. Ghassemi, "MambaNet: A Hybrid Neural Network for Predicting the NBA Playoffs," Oct. 31, 2022, *arXiv*: arXiv:2210.17060. doi: 10.48550/arXiv.2210.17060.
[17] S. Cai, K. Zheng, G. Chen, H. V. Jagadish, B. C. Ooi, and M. Zhang, "ARM-Net: Adaptive Relation Modeling Network for Structured Data," in *Proceedings of the 2021 International Conference on Management of Data*, in SIGMOD '21. New York, NY, USA: Association for Computing Machinery, Jun. 2021, pp. 207–220. doi: 10.1145/3448016.3457321.
[18] R. Khanmohammadi, S. Saba-Sadiya, S. Esfandiarpour, T. Alhanai, and M. M. Ghassemi, "MambaNet: A Hybrid Neural Network for Predicting the NBA Playoffs," Oct. 31, 2022, *arXiv*: arXiv:2210.17060. doi: 10.48550/arXiv.2210.17060.
[19] X. Bouthillier *et al.*, "Accounting for Variance in Machine Learning Benchmarks," *Proceedings of Machine Learning and Systems*, vol. 3, pp. 747–769, Mar. 2021.
[20] J. Bergstra and Y. Bengio, "Random search for hyper-parameter optimization," *J. Mach. Learn. Res.*, vol. 13, no. null, pp. 281–305, Feb. 2012.